\documentclass[letter, 10 pt, conference]{ieeeconf}  

\IEEEoverridecommandlockouts 
\usepackage[english]{babel}
\usepackage[utf8]{inputenc}
\usepackage{times} 
\usepackage{cite}

\usepackage{booktabs}

\usepackage{hyperref}
\usepackage[nolist]{acronym}

\usepackage{multirow}

\usepackage{siunitx}
\usepackage{graphicx} 
\usepackage{epsfig} 
\usepackage{pgfplots}
\usepackage{caption}
\usepackage[font=footnotesize]{subcaption}
\usepackage{pgfplots}

\usepackage{float}

\usepackage[export]{adjustbox}

\usepackage{xcolor}
\definecolor{myYellow}{rgb}{0.93,0.69,0.13}
\definecolor{myPurple}{rgb}{0.49,0.18,0.56}
\definecolor{myGreen}{rgb}{0.26 0.72 0.54}
\definecolor{darkgreen}{rgb}{0.272, 0.50, 0.376}
\definecolor{lightgreen}{rgb}{0.585, 0.82, 0.647}

\colorlet{mydarkblue}{blue!30!black}

\usepackage{mathtools}
\usepackage{nicefrac}
\usepackage{amsmath}
\usepackage{amssymb}
\usepackage{bm} 
\usepackage{scalerel} 

\DeclareMathAlphabet{\pazocal}{OMS}{zplm}{m}{n}

\def\zz{\mathsf{z}}

\DeclareMathOperator*{\minimize}{minimize}

\usepackage{etoolbox}
\makeatletter%
\AfterPreamble{%
	\usepackage{hyperref}%
	\let\oldhypertarget\hypertarget%
	\renewcommand{\hypertarget}[2]{%
		\oldhypertarget{#1}{#2}%
		\protected@write\@mainaux{}{%
			\string\expandafter\string\gdef%
			\string\csname\string\detokenize{#1}\string\endcsname{#2}%
		}%
	}%
	\newcommand{\myhyperlink}[1]{%
		\hyperlink{#1}{\csname #1\endcsname}%
	}%
}
\makeatother%

\newcounter{Remark}
\newcounter{Problem}
\usepackage{algorithm}
\usepackage[noend]{algpseudocode}

\makeatletter
\def\BState{\State\hskip-\ALG@thistlm}
\makeatother

\usepackage{tikz}
\pgfdeclarelayer{foreground}
\pgfsetlayers{background, main, foreground}
\usetikzlibrary{quotes, angles, backgrounds, arrows, automata, shapes, positioning, calc, through, spy, decorations.pathreplacing, decorations.markings, arrows.meta, automata, petri, shapes.multipart}

\tikzset{
    imglabel/.style={
      rectangle,
      inner sep=2pt,
      text=black,
      minimum height=1em,
      text centered,
      fill=white,
      fill opacity=1.0,
      text opacity=1,
      anchor=south west,
    },
  }
\tikzset{
	state/.style={
		rectangle,
		draw=black, very thick,
		minimum height=1.0em,
		text centered,
	},
}
\tikzset{
  on each segment/.style={
    decorate,
    decoration={
      show path construction,
      moveto code={},
      lineto code={
        \path [#1]
        (\tikzinputsegmentfirst) -- (\tikzinputsegmentlast);
      },
      curveto code={
        \path [#1] (\tikzinputsegmentfirst)
        .. controls
        (\tikzinputsegmentsupporta) and (\tikzinputsegmentsupportb)
        ..
        (\tikzinputsegmentlast);
      },
      closepath code={
        \path [#1]
        (\tikzinputsegmentfirst) -- (\tikzinputsegmentlast);
      },
    },
  },
  mid arrow/.style={postaction={decorate,decoration={
        markings,
        mark=at position .5 with {\arrow[#1]{stealth}}
      }}},
}

\tikzset{
  half circle/.style={
      semicircle,
      shape border rotate=180,
      anchor=chord center,
      minimum size=5mm
      }
}

\graphicspath{{./figures/}}

\newcommand\copyrighttext{%
    \small \begin{center} \color{red} \textcopyright\,2025 IEEE. Accepted for presentation to the ``2025 IEEE International Conference on Systems, Man, and Cybernetics (SMC)", 5–8 October 2025, Vienna, Austria. Personal use of this material is permitted. Permission from IEEE must be obtained for all other uses, in any current or future media, including reprinting/republishing this material for advertising or promotional purposes, creating new collective works, for resale or redistribution to servers or lists, or reuse of any copyrighted component of this work in other works. \end{center}}
\newcommand\copyrightnotice{%
	\begin{tikzpicture}[remember picture,overlay]
	\node[anchor=south,yshift=25.6cm] at (current page.south) 
	{\color{red}\fbox{\parbox{\dimexpr\textwidth-\fboxsep-\fboxrule\relax}{\copyrighttext}}};
	\end{tikzpicture}%
}

\title{\copyrightnotice \LARGE \bf Free-Space Optical Communication-Driven NMPC Framework for Multi-Rotor Aerial Vehicles in Structured Inspection Scenarios} 

\author{Giuseppe Silano$^{1,2}$, Daniel Bonilla Licea$^{3,2}$, Hajar  El Hammouti$^{3}$, and Martin Saska$^{2}$  
    \thanks{This work was partially funded by the research fund for the Italian Electrical System under decree n. 388 of November 6th, 2024, by the CTU grant no. SGS23/177/OHK3/3T/13, by the EU under ROBOPROX reg. no. CZ.02.01.01/00/22\_008/0004590, and by the GAČR project no. 23-07517S.}
    \thanks{$^1$Department of Power Generation Technologies and Materials, Ricerca sul Sistema Energetico S.p.A., Milan, Italy. $^2$Department of Cybernetics, Czech Technical University, Prague, Czechia (e-mails: {\tt\small \{silangiu, martin.saska\}@cvut.cz}). $^3$College of Computing, Mohammed VI Polytechnic University, Ben Guerir, Morocco (e-mails: {\tt\small \{daniel.bonilla, hajar.elhammouti\}@um6p.ma)}.}
}

\begin{document}

\maketitle
\thispagestyle{empty} 
\pagestyle{empty} 


\begin{acronym}
    \acro{BER}[BER]{Bit Error Rate}
    \acro{BS}[BS]{Base Station}
    \acro{CCW}[CCW]{Counter-ClockWise}
    \acro{CW}[CW]{ClockWise}
    \acro{CoM}[CoM]{Center of Mass}
    \acro{FoV}[FoV]{Field-of-View}
    \acro{FSO}[FSO]{Free-Space Optical}
    \acro{GTMR}[GTMR]{Generically-Tilted Multi-Rotor}
    \acro{LED}[LED]{Light-Emitting Diode}
    \acro{LoS}[LoS]{Line-of-Sight}
    \acro{MPC}[MPC]{Model Predictive Control}
    \acro{MRAV}[MRAV]{Multi-Rotor Aerial Vehicle}
    \acro{NLP}[NLP]{Nonlinear Programming}
    \acro{NMPC}[NMPC]{Nonlinear Model Predictive Control}
    \acro{OW}[OW]{Optical Wireless}
    \acro{PAT}[PAT]{Pointing, Acquisition and Tracking}
    \acro{QoS}[QoS]{Quality of Service}
    \acro{QP}[QP]{Quadratic Programming}
    \acro{RF}[RF]{Radio-Frequency}
    \acro{ROS}[ROS]{Robot Operating System}
    \acro{SNR}[SNR]{Signal-to-Noise-Ratio}
    \acro{UAV}[UAV]{Unmanned Aerial Vehicle}
    \acro{UGV}[UGV]{Unmanned Ground Vehicle}
    \acro{wrt}[w.r.t.]{with respect to}
\end{acronym}



\begin{abstract}

    This paper introduces a \ac{NMPC} framework for communication-aware motion planning of \acp{MRAV} using \ac{FSO} links. The scenario involves \acp{MRAV} equipped with body-fixed optical transmitters and \acp{UGV} acting as mobile relays, each outfitted with fixed conical \ac{FoV} receivers. The controller integrates optical connectivity constraints into the \ac{NMPC} formulation to ensure beam alignment and minimum link quality, while also enabling \ac{UGV} tracking and obstacle avoidance. The method supports both coplanar and tilted \ac{MRAV} configurations. MATLAB simulations demonstrate its feasibility and effectiveness.

\end{abstract}



%
%



\section{Introduction}
\label{sec:introduction}

Ensuring the reliability of energy infrastructure is essential
for the resilience of modern power systems. As electric grids grow more complex, with increased renewable integration and bidirectional flows, frequent inspection of high-voltage components becomes necessary to prevent failures and guarantee service continuity \cite{iea2021grids}. Traditional inspection methods are labor-intensive, costly (\$2,000/hour), and potentially hazardous, especially in remote or constrained areas \cite{Ollero2024BookChapter}. 

\acfp{MRAV} offer a scalable and safer alternative, enabling close-range visual inspection while reducing cost and risk \cite{AhmedJINT2024, Caballero2023IEEEAccess, Silano2025RAS}. 
However, their use in large-scale deployments introduces communication challenges. High-bandwidth, low-latency data transmission is crucial to support real-time diagnostics by remote operators \cite{AhmedJINT2024}. Conventional \ac{RF} links suffer from limited bandwidth, congestion, and susceptibility to interference and eavesdropping \cite{LiceaComMag2025}. \acf{FSO} communication is an attractive complement, offering secure, high-rate links via directional light beams. However, \ac{FSO} systems are sensitive to beam misalignment, range, and environmental interference \cite{KhalighiIEEEComSurTut2014}. A stable link requires the transmitter to remain within the \acf{FoV} of the receiver and aligned within a narrow angular tolerance (see Figure~\ref{fig:connectivityOpticalSystem}).

\begin{figure}[tb]
    \centering
    \includegraphics[width=0.85\columnwidth]{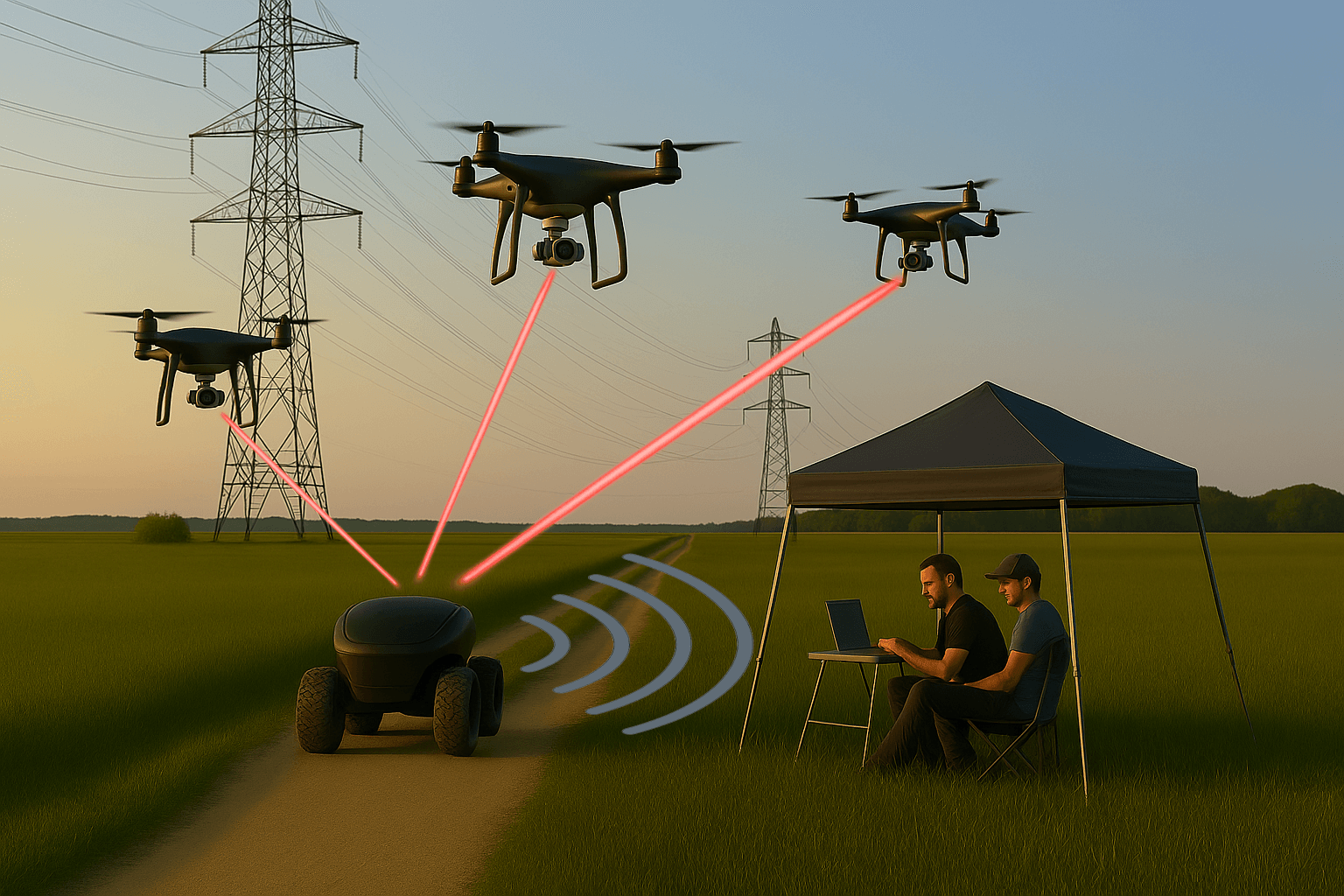}
    \vspace{-0.5em}
    \caption{Illustration of the proposed inspection scenario with \ac{MRAV}–\ac{UGV} communication to a remote base station.}
    \label{fig:scenario}
    \vspace*{-1.5em}
\end{figure}

In this context, we consider a heterogeneous robotic system for infrastructure inspection, where \acp{MRAV} carry rigidly mounted optical transmitters, and \acfp{UGV} act as mobile optical relays, forwarding data to a remote base station via \ac{RF} links, as illustrated in Figure~\ref{fig:scenario}. Each \ac{UGV} is equipped with multiple actuated receivers featuring wide conical \acp{FoV}. The \ac{UGV} handles beam acquisition via local actuation, while tracking and alignment must be maintained by the \ac{MRAV} through motion and attitude control. 

This scenario poses coupled geometric and communication constraints: the \ac{MRAV} must track a moving \ac{UGV}, keep the optical beam within the acceptance cone of the assigned receiver, and maintain a valid communication range, all while respecting actuation limits and avoiding collisions. The problem is further complicated by the need to operate in cluttered environments and under dynamic conditions.

While \ac{FSO} communication has been explored in underwater and aerial networks \cite{Hoeher2021ComSurvey, Li2024IEEETransVehTech}, its use in terrestrial mobile robotics remains limited due to alignment sensitivity and \ac{LoS} constraints. Prior robotic \ac{FSO} systems often assume static or low-mobility agents \cite{Li2024IEEETransVehTech, Dabiri2022TWC, DabiriIEEECL2020}, and many communication-aware planning strategies rely on heuristics or \ac{RF}-specific assumptions \cite{Ghaffarkhah2011TAC, Bonilla2024IEEEProc}, lacking integration with vehicle dynamics or beam geometry.

\begin{figure}[tb]
    \centering
    \scalebox{0.8}{
    \begin{tikzpicture}
        \node (drone) at (0,0) [text centered]{
            \adjincludegraphics[width=0.90\columnwidth, trim={{0.\width} {0.\height} {0.\width} {0.\height}}, clip]{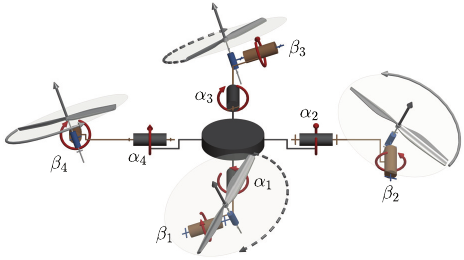}};

            \draw[-latex] (-4,-2) -- (-4,-1) node[left]{$\mathbf{z}_W$} ;
            \draw[-latex] (-4,-2) -- (-3,-2) node[below]{$\mathbf{x}_W$};
            \draw[-latex] (-4,-2) -- (-3.5,-1.5) node[above]{$\mathbf{y}_W$};
            \draw (-4,-2) node[below]{$O_W$};
    
            \draw[-latex, red] (0,0) -- (0,1) node[left]{$\mathbf{z}_B$} ;
            \draw[-latex, red] (0,0) -- (1,0) node[above]{$\mathbf{x}_B$};
            \draw[-latex, red] (0,0) -- (0.5,0.5) node[above]{$\mathbf{y}_B$};
            \draw[red] (0,0) node[above left]{$O_B$};
    
            \draw[dashed, red] (-4,-2) -- node[above]{$\mathbf{p}$} (0,0);
            \draw[-latex, red] (-4,-2) to [out=5, in=-70] node[above]{$\mathbf{R}$} (0,0);
    
            \node (sensor) at (2.1,-2.6) [text centered, rotate=-10]{
                \adjincludegraphics[width=0.11\columnwidth, trim={{0.\width} {0.\height} {0.\width} {0.\height}}, clip]{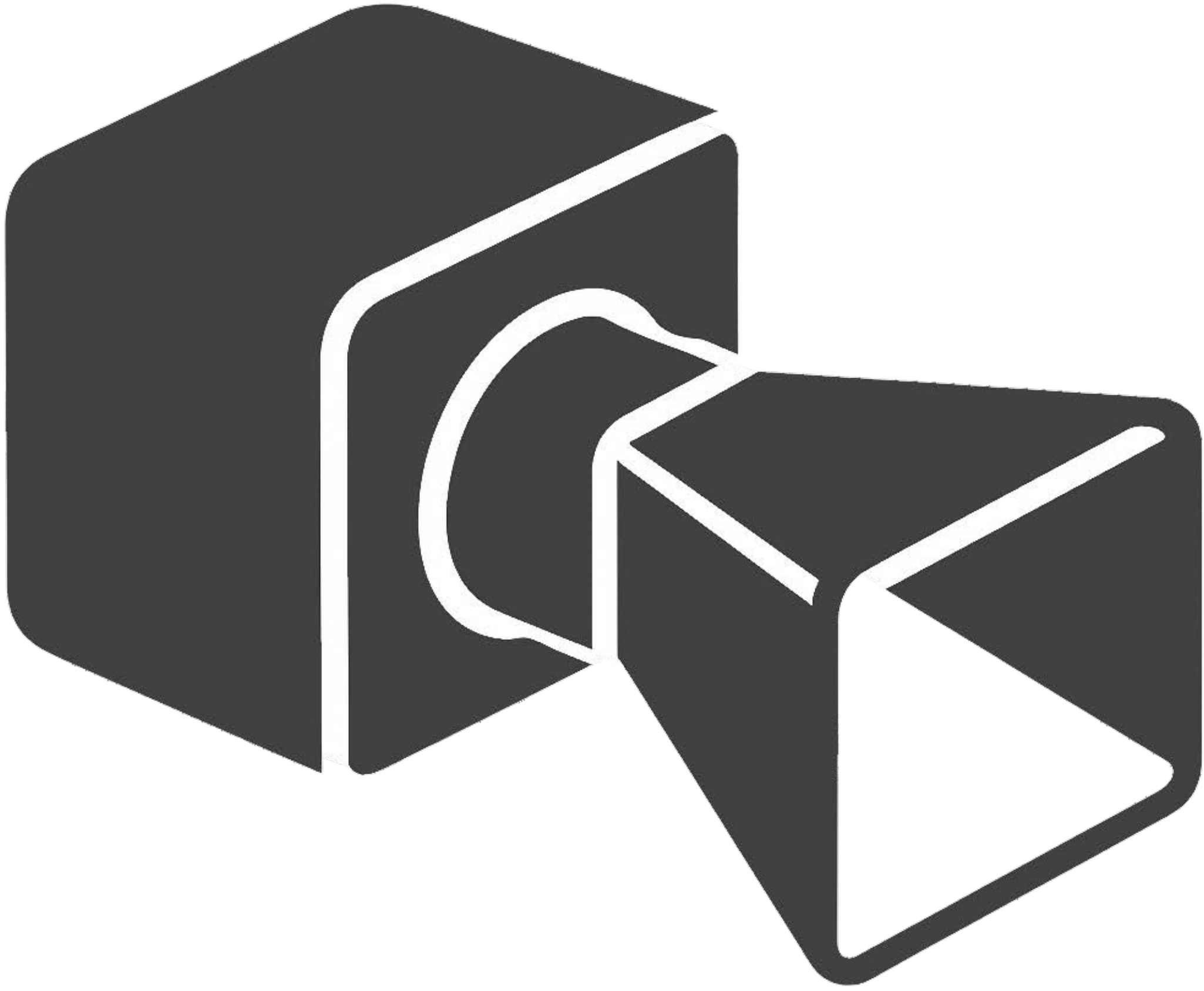}};
    
            \draw[latex-] (2.0,-3.5) node[left]{$\mathbf{y}_T$} -- (2.0,-2.5);
            \draw[-latex] (2.0,-2.5) -- (2.65,-2.95) node[below]{$\mathbf{z}_T$};
            \draw[latex-] (1.5,-3.0) node[above left]{$\mathbf{x}_T$} -- (2.0,-2.5) ;
            \draw (2.35,-2.7) node[above]{$O_T$};
    
            \draw[dashed, red] (0,0) -- node[below, yshift=-1.5ex]{$\mathbf{p}_\mathrm{BT}$} (2.0,-2.5);
            \draw[-latex, red] (0,0) to [out=5, in=70] node[below, yshift=-2.5ex, xshift=5ex]{$\mathbf{R}_\mathrm{BT}$} (2.0,-2.5);
        
    \end{tikzpicture}    
    }
    \vspace*{-1em}
    \caption{Quadrotor with optical transmitter. Zero tilt angles $\{\alpha_i$,  $\beta_i\}$ define a coplanar setup; nonzero values define a tilted one \cite{HamandiIJRR2021}.} 
    \label{fig:platform}
    \vspace*{-1em}
\end{figure}
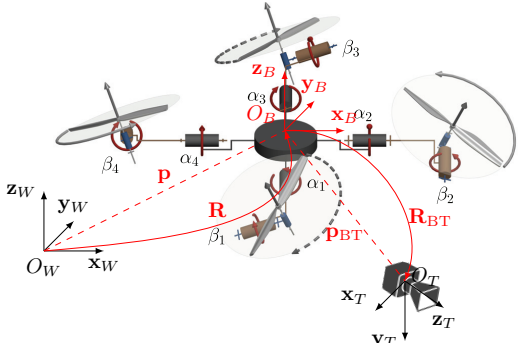

\ac{MPC} offers a principled approach to trajectory planning under complex constraints by explicitly incorporating system dynamics, actuation limits, and predictive planning within a unified optimization problem. Recent \ac{MPC}-based methods have addressed \ac{RF} connectivity \cite{Liu2020TCNS, JafarzadehITSC2021} or \ac{FoV}-based perception \cite{Falanga2018IROS, Penin2018RAL}, but do not generalize to directional and alignment-sensitive \ac{FSO} links. These links introduce stronger spatial coupling, requiring precise control over beam orientation, platform motion, and inter-agent distance.

To address this gap, this paper proposes a Nonlinear \ac{MPC} (\acs{NMPC}) framework for communication-aware motion planning in aerial-ground systems with directional optical links. The \ac{MRAV} is modeled as a \acf{GTMR} \cite{Ryll2019IJRR, HamandiIJRR2021}, enabling applicability to both coplanar and tilted architectures. Tilted platforms offer enhanced maneuverability, while coplanar designs remain common for their simplicity and widespread use. Figure~\ref{fig:platform} illustrates these configurations. The \acs{NMPC} formulation embeds beam alignment and range constraints as hard requirements and uses soft penalties for obstacle avoidance. The result is a unified control strategy that enforces communication reliability, tracking accuracy, and safety. MATLAB simulations validate the approach in a power infrastructure inspection scenario, demonstrating consistent beam alignment, reliable link quality, and feasibility under realistic operating conditions.



\vspace*{-1em}
\section{System Modeling}
\label{sec:systemModeling}

This section describes the dynamic model of the \ac{MRAV}, the optical transceiver geometry, and the communication constraints. Table~\ref{tab:tableOfNotation} summarizes the notation.

\begin{table}[tb]
    \centering
    \caption{Notation summary.}
    \vspace{-0.5em}
    \label{tab:tableOfNotation}
    \begin{adjustbox}{max width=\columnwidth}
    \begin{tabular}{p{2.5cm} p{5.5cm}}
        \toprule
        $\mathcal{F}_W$, $\mathcal{F}_B$, $\mathcal{F}_T$, $\mathcal{F}_R$ & World, body, transmitter, and receiver frames \\
        $\mathbf{p}$, $\mathbf{v}$, $\bm{\eta}$, $\bm{\omega}$, $\bm{\xi}$, $\bm{\tau}$ & Pos., vel., orient., thrust, and torque (\ac{MRAV}) \\
        $c_\xi$, $c_\tau$, $\mathbf{p}_m$, $\zz_P$, $\Omega$ & Thrust/torque coeff., motor pos., axis, rotor speed squared \\
        $N_p$, $\mathbf{J}$, $\varphi$, $\vartheta$, $\psi$ & Propeller count, inertia, roll, pitch, yaw \\
        $\mathbf{t}$, $T_s$, $N$, $\mathbf{x}$, $\mathbf{u}$, $\alpha$, $\beta$ & Time horizon, step, state/control, tilt angles \\
        $\mathbf{R}$, $\mathbf{T}$, $\mathbf{F}$, $\mathbf{M}$ & Rotation and allocation matrices \\
        $g$, $m$, $\mathbf{e}_3$, $\bar{\mathbf{x}}$, $\bar{\mathbf{u}}$ & Gravity, mass, unit vector, extended state/control \\
        $\underline{\gamma}$, $\overline{\gamma}$, $\dot{\underline{\gamma}}$, $\dot{\bar{\gamma}}$ & Rotor speed/acceleration bounds \\
        $\mathbf{p}_\mathrm{BT}$, $\mathbf{R}_\mathrm{BT}$ & Transmitter offset and rotation (in body frame) \\
        $\psi_C$, $\mathbf{d}_C$, $\mathbf{p}_R$, $\Phi_{1/2}$ & Cone angle, link vector, receiver pos., half-power bandwith \\
        $\underline{d}_\mathcal{C}$, $\overline{d}_\mathcal{C}$, $I_\mathrm{tx}$, $I_\mathrm{rx}$, $T_\mathrm{rx}$ & Link range bounds, binary flags, time window \\
        $I$, $\bar{I}$, $c_\delta$, $\dot{c}_\delta$ & Connectivity metrics, mis. cosine and derivative \\
        $\mathbf{p}_\mathcal{O}$, $d_\mathcal{O}$, $N_\mathcal{O}$, $d_\mathrm{safe}$  & Obs. pos., size, count, safety margin \\
        $\bar{\mathbf{x}}_d$, $\mathbf{y}_d$, $\mathbf{y}$, $\Upsilon$ & Reference state/output, desired link distance \\
        $\delta$, $\bar{\mathbf{Q}}$, $\mathbf{Q}$, $\mathbf{Q}_{\bar{\mathbf{u}}}$, $\mathbf{Q}_{\bm{\varepsilon}}$, $\varepsilon$ & Misalignment angle, cost weights,  slack var. \\
        \bottomrule
    \end{tabular}
    \end{adjustbox}
    \vspace*{-2em}
\end{table}



\subsection{Multi-rotor dynamics}
\label{sec:multiRotorDynamics}

The \ac{MRAV} is modeled as a \ac{GTMR} system \cite{Ryll2019IJRR, HamandiIJRR2021}, a flexible framework that accommodates both coplanar (under-actuated) and tilted (fully-actuated) configurations. The discrete-time dynamics $x_{k+1} = f(x_k, u_k)$, where $x_k \in \pazocal{X} \subset \mathbb{R}^n$ is the system state and $u_k \in \pazocal{U} \subset \mathbb{R}^m$ is the control input at time step $k$. The vehicle is actuated by $N_p$ motor-propeller pairs, each arbitrarily positioned and oriented \ac{wrt} the body frame, as depicted in Figure \ref{fig:platform}. 

The motion is described \ac{wrt} to the world frame $\pazocal{F}_W = \{O_W, \mathbf{x}_W, \mathbf{y}_W, \mathbf{z}_W\}$ and the body-fixed frame $\pazocal{F}_B = \{O_B, \mathbf{x}_B, \mathbf{y}_B, \mathbf{z}_B\}$, with origin at the \ac{CoM}. The state vector includes position $\mathbf{p} \in \mathbb{R}^3$, velocity $\mathbf{v} = \dot{\mathbf{p}}$ in $\pazocal{F}_W$, orientation $\bm{\eta} = (\varphi, \vartheta, \psi)^\top \in \mathbb{R}^3$ as Euler angles, and angular velocity $\bm{\omega} \in \mathbb{R}^3$ in $\pazocal{F}_B$. The control input $\bm{\xi} \in \mathbb{R}^{N_p}$ collects the thrusts contributions from each rotor. Each rotor $i \in \{1, \dots, N_p\}$ generates a thrust force $\xi_i$ and torque $\tau_i$, defined as $\xi_i = c_{\xi_i} \Omega_i \zz_{P_i}$ and $\tau_i = \left(c_{\xi_i} \mathbf{p}_{m_i} \times \zz_{P_i} + c_{\tau_i} \zz_{P_i}\right) \Omega_i$, where $\Omega_i$ is the squared rotor speed, $c_{\xi_i}$ and $c_{\tau_i}$ are thrust and torque coefficients,  $\mathbf{p}_{m_i}$ is the motor's position in $\pazocal{F}_B$, and $\zz_{P_i} \in \mathbb{S}^2$ is its rotation axis \cite{Ryll2019IJRR}.

Let $\mathbf{t} = (t_0, \dots, t_N)^\top \in \mathbb{R}^{N+1}$ be the discrete time horizon with step size $T_s > 0$. The state and control sequences are $\mathbf{x} = (\mathbf{p}^\top, \bm{\eta}^\top, \mathbf{v}^\top, \bm{\omega}^\top)^\top \in \mathbb{R}^{13 \times N}$ and $\mathbf{u} = \bm{\xi} \in \mathbb{R}^{N_p \times N}$, with $\bullet_k$ denoting the value at step $k$.

The continuous-time dynamics are modeled using the Newton–Euler formulation: 
\begin{equation}\label{eq:multirotorDynamics}
\left\{
    \begin{array}{l}
    \dot{\mathbf{p}} = \mathbf{v} \\
    \dot{\bm{\eta}} = \mathbf{T}(\bm{\eta}) \bm{\omega} \\
    m \dot{\mathbf{v}} = -mg \mathbf{e}_3 + \mathbf{R}(\bm{\eta}) \mathbf{F}(\bm{\alpha}, \bm{\beta}) \mathbf{u} \\
    \mathbf{J} \dot{\bm{\omega}} = - \bm{\omega} \times \mathbf{J} \bm{\omega} + \mathbf{M}(\bm{\alpha}, \bm{\beta}) \mathbf{u}
    \end{array}
\right.,
\end{equation}
where $m$ is the vehicle's mass, $g$ is gravity, and $\mathbf{e}_3 = [0,0,1]^\top$. The rotation matrix $\mathbf{R}(\bm{\eta})$ maps vectors from $\pazocal{F}_B$ to $\pazocal{F}_W$, while the matrix $\mathbf{T}(\bm{\eta})$ transforms angular velocity to Euler angle rates. The inertia matrix $\mathbf{J} \in \mathbb{R}^{3 \times 3}$ is symmetric and positive-definite. The force and torque allocation matrices $\mathbf{F}(\bm{\alpha}, \bm{\beta})$ and $\mathbf{M}(\bm{\alpha}, \bm{\beta})$ depend on motor orientation parameters $\{\alpha_i, \beta_i\}$, as depicted in Figure~\ref{fig:platform} \cite{Ryll2019IJRR}.

To account for actuator dynamics, rotor inputs are extended with a smoothness constraint. Letting $\bar{\mathbf{x}} = (\mathbf{x}^\top, \mathbf{u}^\top)^\top$ and $\bar{\mathbf{u}}=\dot{\mathbf{u}}$, the extended dynamics become $\dot{\mathbf{x}}=f(\bar{\mathbf{x}}, \bar{\mathbf{u}})$. This enables direct enforcement of actuator limits: $\underline{\gamma} \leq \mathbf{u} \leq \bar{\gamma}$ and $\dot{\underline{\gamma}} \leq \bar{\mathbf{u}} \leq \dot{\bar{\gamma}}$, where the bounds define allowable rotor speeds and accelerations. Enforcing these constraints ensure physically valid thrust profiles and support closed-loop \ac{NMPC} stability under agile maneuvers or precise tracking.



\subsection{Transceiver model}
\label{sec:transceiverModel}

To enable directional communication, the \ac{MRAV} is modeled as carrying a rigidly mounted optical transmitter, such as a laser diode or LED source. The transmitter is aligned with a local reference frame $\mathcal{F}_T = \{O_T, \mathbf{x}_T, \mathbf{y}_T, \mathbf{z}_T\}$, with the beam axis pointing along $\mathbf{z}_T$. Its fixed position and orientation in the body frame $\mathcal{F}_B$ are given by $\mathbf{p}_\mathrm{BT} \in \mathbb{R}^3$ and rotation matrix $\mathbf{R}_\mathrm{BT} \in \text{SO}(3)$, respectively. This sensor-agnostic model allows the framework to accommodate a range of emitter types and mounting configurations. Figure~\ref{fig:platform} illustrates this configuration.

On the ground, each \ac{UGV} is equipped with one or more photodiode-based receivers, each defining a conical \ac{FoV} with aperture angle $\psi_\mathcal{C}$, centered along the unit vector $\mathbf{z}_R$ in its own frame $\mathcal{F}_R = \{O_R, \mathbf{x}_R, \mathbf{y}_R, \mathbf{z}_R\}$. The cone represents the region in which the transmitter must lie for a valid link to be established. The \ac{UGV} is assumed to have the ability to steer its receiver axis to track the aerial transmitter, enabling dynamic alignment. An illustration of this configuration is shown in Figure~\ref{fig:connectivityOpticalSystem}. 

Let $\mathbf{p}_R$ be the position of the assigned receiver and $\mathbf{p}$ the position of the \ac{MRAV}, both expressed in $\pazocal{F}_W$. The relative position vector is $\mathbf{d}_\mathcal{C} = \mathbf{p} - \mathbf{p}_R$. For a valid optical link, this vector must satisfy both angular and distance constraints: the beam emitted by the transmitter must lie within the receiver’s acceptance cone and remain within an admissible range. These requirements are enforced by requiring the alignment condition $\nicefrac{\mathbf{z}_T^\top (-\mathbf{d}_\mathcal{C})}{\|\mathbf{d}_\mathcal{C}\|} \geq \cos(\psi_\mathcal{C})$ and the distance constraint $\underline{d}_\mathcal{C} \leq \|\mathbf{d}_\mathcal{C}\| \leq \overline{d}_\mathcal{C}$. The first ensures proper beam–cone alignment, while the second bounds signal range: $\overline{d}_\mathcal{C}$ is set by the power budget to sustain the target bit rate \cite{KhalighiIEEEComSurTut2014}, and $\underline{d}_\mathcal{C}$ prevents saturation or geometric misalignment. Both are time-varying and treated as hard constraints in the \ac{NMPC} formulation.



\subsection{Optical communication system}
\label{sec:opticalCommunicationSystem}

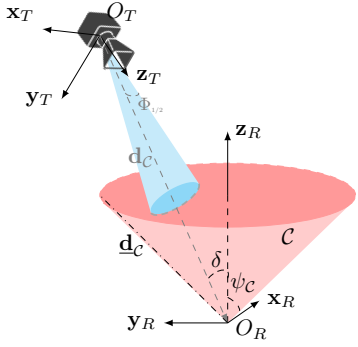
\begin{figure}[tb]
    \centering
    \scalebox{0.85}{
    \begin{tikzpicture}
        
        \newcommand{\radiusxTransmitter}{0.5}
        \newcommand{\radiusyTransmitter}{0.20}
        \newcommand{\heightTransmitter}{-2}
        \coordinate (a) at (-{\radiusxTransmitter*sqrt(1-(\radiusyTransmitter/\heightTransmitter)*(\radiusyTransmitter/\heightTransmitter))},{\radiusyTransmitter*(\radiusyTransmitter/\heightTransmitter)});
        \coordinate (b) at ({\radiusxTransmitter*sqrt(1-(\radiusyTransmitter/\heightTransmitter)*(\radiusyTransmitter/\heightTransmitter))},{\radiusyTransmitter*(\radiusyTransmitter/\heightTransmitter)});

        

        \newcommand{\radiusx}{2}
        \newcommand{\radiusy}{.5}
        \newcommand{\height}{-2}
        \coordinate (a) at (-{\radiusx*sqrt(1-(\radiusy/\height)*(\radiusy/\height))},{\radiusy*(\radiusy/\height)});
        \coordinate (b) at ({\radiusx*sqrt(1-(\radiusy/\height)*(\radiusy/\height))},{\radiusy*(\radiusy/\height)});
        
        \fill[red!20] (a)--(0,\height)--(b)--cycle;
        \fill[red!40] circle (\radiusx{} and \radiusy);
        
        \begin{scope}
            \clip ([xshift=-2mm]a) rectangle ($(b)+(1mm,-2*\radiusy)$);
            \draw[red!40] circle (\radiusx{} and \radiusy);
        \end{scope}
        \begin{scope}
            \clip ([xshift=-2mm]a) rectangle ($(b)+(1mm,2*\radiusy)$);
            \draw[dashed, red!40] circle (\radiusx{} and \radiusy);
        \end{scope}

        \fill[fill=cyan!20] (-1.2,-0.28) -- (-2.0,2.5) -- (-0.45,0.18) -- cycle;
        \fill[fill=cyan!40, xshift=-0.1em, yshift=-0.15em, rotate=-60] (-0.4,-0.68) circle (0.15 and 0.43);
        \draw[gray!70, densely dashed, line width=0.75pt] (-1.2,-0.28) to [controls=+(-50:.2) and +(-70:.2)] (-0.45,0.18);
        

        \draw[gray, densely dashed] (-1.60,1.6) to[out=-70,in=-40] (-1.40,1.6);
        \draw[gray] (-1.2,1.15) node[above]{\footnotesize $\Phi_{\scaleto{1/2}{0.35em}}$};


            
        \node (sensor) at (-1.9,2.4) [text centered, rotate=-30]{
            \adjincludegraphics[width=0.11\columnwidth, trim={{0.\width} {0.\height} {0.\width} {0.\height}}, clip]{figure/sensor.pdf}};
        
        \draw[latex-] (-2.6,1.5) node[left]{$\mathbf{y}_T$} -- (-2.0,2.5);
        \draw[-latex] (-2.0,2.5) -- (-1.55,1.85) node[right]{$\mathbf{z}_T$};
        \draw[latex-] (-2.9,2.6) node[above left]{$\mathbf{x}_T$} -- (-2.0,2.5);
        \draw (-1.65,2.6) node[above]{$O_T$};

        \draw[dashed] (0,\height) -- (0,0);
        \draw[-latex] (0,0) -- (0,1) node[right]{$\mathbf{z}_R$};
        \draw[latex-] (-1,\height) node[left]{$\mathbf{y}_R$} -- (0,\height);
        \draw[-latex] (0,\height) -- (0.5,\height+0.35) node[right]{$\mathbf{x}_R$};
        \draw (0,\height-0.1) node[right]{$O_R$};

        \draw[gray, dashed, -latex] (-2.0,2.5) --  node[above, xshift=-1.0em, yshift=0.2em]{$\mathbf{d}_\mathcal{C}$} (0,\height);
        \draw[dash dot] (0,\height) -- node[above, xshift=-1.4em, yshift=-0.2em]{$\underline{\mathbf{d}}_\mathcal{C}$} (-\radiusx,0);
        
        
        \draw (0.25,-1.65) node[above]{$\psi_\mathcal{C}$};
        \draw[dashed] (0,-1.6) arc (25:0:2cm and 0.5cm); 
        \draw[densely dashed] (-0.3,-1.3) to [controls=+(50:.2) and +(70:.2)]  (0,-1.3);
        \draw (-0.175,-1.25) node[above]{$\delta$};
        
        \node(coneText) at (0.95,-0.65) [text centered]{$\mathcal{C}$}; 

    \end{tikzpicture}
    }
    \vspace{-0.7em}
    \caption{Transceiver geometry with idealized receiver cone. Misalignment angle $\delta$ (see Section \ref{sec:objectiveFunction}) quantifies the deviation between  beam direction $\mathbf{z}_T$ and receiver axis $\mathbf{z}_R$.}
    \label{fig:connectivityOpticalSystem}
    \vspace*{-1.5em}
\end{figure}

To determine the validity of an optical link, we define a binary indicator $I$ based on beam–receiver alignment. The directional beam emitted by the transmitter has a central lobe characterized by a half-power beamwidth $\Phi_{1/2}$, while the receiver has a wider conical \ac{FoV} with aperture $\psi_\mathcal{C}$ (typically $\psi_\mathcal{C} \gg \Phi_{1/2}$), as depicted in Figure~\ref{fig:connectivityOpticalSystem}. The link status is defined as $I=I_\mathrm{tx}I_\mathrm{rx}$, where the transmitter flag $I_\mathrm{tx}$ equals $1$ if $\nicefrac{\mathbf{z}_T^\top (-\mathbf{d}_\mathcal{C})}{\|\mathbf{d}_\mathcal{C}\|} \geq \cos(\Phi_{1/2})$, and $0$ otherwise. Similarly, the receiver flag $I_\mathrm{rx}$ equals $1$ if $\nicefrac{\mathbf{z}_R^\top (\mathbf{d}_\mathcal{C})}{\|\mathbf{d}_\mathcal{C}\|} \geq \cos(\psi_\mathcal{C})$, and $0$ otherwise. 

The indicators $I_\mathrm{tx}$ and $I_\mathrm{rx}$ encode directional alignment: $I_\mathrm{tx} = 1$ if the receiver lies within the main lobe of the transmitter beam, and $I_\mathrm{rx} = 1$ if the transmitter is inside the receiver's acceptance cone. The condition $I_\mathrm{tx} = 1$ implies precise beam alignment within the transmitter's angular footprint. Given the beam's narrow profile, partial edge overlap is insufficient, as optical power decays rapidly outside the main lobe. This reflects the system's high directional selectivity.

To assess temporal link quality, we define a moving average  $\bar{I}(t_k) = \frac{1}{T_\mathrm{rx}} \int_{t_k - T_\mathrm{rx}}^{t_k} I(\zeta), \mathrm{d}\zeta$, where $\bar{I} \in [0, 1]$ measures the fraction of time the link is maintained over a window of duration $T_\mathrm{rx}$. High values of $\bar{I}$ are critical for real-time data streams, while lower values may suffice in buffered or delay-tolerant applications. Nonetheless, maximizing $\bar{I}$ is beneficial in cluttered or dynamic environments. This formulation provides a physically grounded and computationally tractable representation of \ac{FSO} connectivity, suitable for integration into a predictive control framework.



\section{Optimal Control Problem Formulation}
\label{sec:optimalControlProblemFormulation}

The goal is to control a \ac{MRAV} with a fixed optical transmitter to track a moving \ac{UGV}, maintain beam alignment within the receiver's acceptance cone, and avoid collisions, all while satisfying actuation and communication constraints. The controller guarantees link reliability and system feasibility over the prediction horizon. The trajectories of the \ac{UGV} and all relevant static and dynamic obstacles are assumed known, as is realistic in structured tasks like power grid inspection, where maps are pre-surveyed and dynamic agents (e.g., personnel, vehicles) follow predictable paths. The \ac{UGV} path can be broadcast online via a coordination layer \cite{Cantieri2020Sensors}.



\subsection{Collision avoidance}
\label{sec:collisionAvoidance}

To ensure safety during inspection missions, the proposed \ac{NMPC} framework incorporates a predictive collision avoidance strategy that accounts for both static and dynamic obstacles. Each obstacle is modeled as a closed ball $\mathcal{O}_j \subset \mathbb{R}^3$, centered at position $\mathbf{p}_{\mathcal{O}_j} \in \mathbb{R}^3$ and radius $d_{\mathcal{O}_j} > 0$. Given $\mathbf{p}$ as the \ac{MRAV}'s position, a minimum separation distance $d_\mathrm{safe} > 0$ is enforced to maintain a safety buffer from each obstacle, requiring $\lVert \mathbf{p} - \mathbf{p}_{{\mathcal{O}_j}} \rVert^2 \geq d_{\mathcal{O}_j} + d_\mathrm{safe}$, $\forall j \in \{1, \dots, N_\mathcal{O}\}$, where $N_\mathcal{O}$ is the total number of obstacles considered in the scene. To preserve the problem feasibility in environments where obstacles are densely packed or dynamically evolving, the hard constraint is relaxed using non-negative slack variable $\varepsilon_j \geq 0$, one for each obstacle. The relaxed constraint becomes $\lVert \mathbf{p} - \mathbf{p}_{\mathcal{O}_j} \rVert^2 \geq d_{\mathcal{O}_j} + d_\mathrm{safe} - \varepsilon_j$.

These slack variables allow the optimization to tolerate limited constraint violations, preventing infeasibility in edge cases while still penalizing unsafe proximity through the cost function. A penalty term $\lVert \bm{\varepsilon} \rVert^2_{\mathbf{Q}_\varepsilon}$ is added to the objective, where $\bm{\varepsilon} = (\varepsilon_1, \dots, \varepsilon_{N_\mathcal{O}})^\top$ and $\mathbf{Q}_\varepsilon$ is a diagonal positive-definite matrix that scales the cost of each violation.

The position of each obstacle $\mathbf{p}_{\mathcal{O}_j}$ is assumed to be known or predictable over the planning horizon. As said, this is a realistic assumption in structured inspection scenarios such as power line monitoring, where static obstacles (e.g., towers, poles, vegetation) are pre-mapped and mobile elements (e.g., ground vehicles or workers) follow known trajectories or broadcast their position periodically \cite{Cantieri2020Sensors}. This predictive knowledge enables the controller to plan ahead, avoiding reactive or overly conservative behavior.



\subsection{Objective function}
\label{sec:objectiveFunction}

The control objective is to ensure that the \ac{MRAV} tracks a reference trajectory provided by the \ac{UGV}, while maintaining beam alignment for uninterrupted optical communication and avoiding obstacles. The motion reference includes position, velocity, and acceleration, encapsulated in a desired state vector $\bar{\mathbf{x}}_d = (\mathbf{p}_d^\top, \mathbf{v}_d^\top, \dot{\mathbf{v}}_d^\top)^\top$, and can be either precomputed or transmitted online via a coordination interface \cite{Cantieri2020Sensors}.

Tracking is formulated as a quadratic cost on the deviation from this reference, defined as $\lVert \bar{\mathbf{x}} - \bar{\mathbf{x}}_d \rVert^2_{\bar{\mathbf{Q}}}$, where $\bar{\mathbf{Q}}$ is a diagonal positive-definite weight matrix encoding the relative importance of each term in the state vector. This ensures the \ac{MRAV} closely follows the intended inspection path in terms of both kinematics and dynamics.

To ensure reliable communication, the objective function penalizes beam misalignment and deviations from the desired link distance. Beam–receiver alignment is measured by the angle $\delta = \arccos\left( \nicefrac{\mathbf{z}_T^\top (-\mathbf{d}_\mathcal{C})}{\lVert \mathbf{d}_\mathcal{C} \rVert} \right)$, where $\mathbf{z}_T$ is the beam direction and $\mathbf{d}_\mathcal{C}$ is the vector from the receiver to the transmitter. Instead of directly penalizing $\delta$, the cost includes $c_\delta = \cos{(\delta)}$, which offers numerical smoothness and boundedness. Additionally, its derivative $\dot{c}_\delta$ is penalized to discourage jitter and improve link stability.

Communication quality is further reinforced by penalizing deviations from a target communication distance $\Upsilon > 0$. The scalar error $\lVert \mathbf{d}_\mathcal{C} \rVert - \Upsilon$ captures deviations from this nominal range, which could degrade link strength or cause alignment issues.

All tracking, alignment, and distance objectives are compactly represented by the output vector $\mathbf{y} = (\mathbf{p}^\top, \mathbf{v}^\top, \dot{\mathbf{v}}^\top, c_\delta, \dot{c}_\delta, \lVert \mathbf{d}_\mathcal{C} \rVert )^\top$, with a reference $\mathbf{y}_d = (\mathbf{p}_d^\top, \mathbf{v}_d^\top, \dot{\mathbf{v}}_d^\top, 1, 0, \Upsilon)^\top$. The overall tracking cost is expressed as $\lVert \mathbf{y} - \mathbf{y}_d \rVert^2_\mathbf{Q}$, where $\mathbf{Q}$ is a weighting matrix balancing motion and communication objectives. This formulation defines a unified cost function that integrates accurate motion tracking, beam alignment, range maintenance, and safe operation in cluttered environments.



\subsection{Optimal control problem}
\label{sec:optimalControlProblem}

The control framework is posed as a constrained discrete-time \acs{NMPC} problem. At each $t_k = kT_s$, with horizon length $N$, the controller computes the optimal control sequence by solving the finite-horizon optimization:

\vspace*{-1.6em}
\begin{subequations}\label{eq:NMPC_formulation}  
    \small
    \begin{align} 
    \minimize_{\bar{\mathbf{x}}, \bar{\mathbf{u}}, \bm{\varepsilon}} \quad & \sum_{k=0}^{N} \lVert \mathbf{y}_{d,k} - \mathbf{y}_k \rVert^2_{\mathbf{Q}} + \lVert \bar{\mathbf{u}}_k \rVert^2_{\mathbf{Q}_{\bar{\mathbf{u}}}} + \lVert \bm{\varepsilon}_k \rVert^2_{\mathbf{Q}_{\bm{\varepsilon}}} \label{subeq:objectiveFunction} \\ 
    \text{s.t.} \quad & \bar{\mathbf{x}}_0 = \bar{\mathbf{x}}(t_0), \label{subeq:initialCondition} \\ 
    & \bar{\mathbf{x}}_{k+1} = f(\bar{\mathbf{x}}_k, \bar{\mathbf{u}}_k), \; k = \{0,\dots,N-1\}, \label{subeq:systemDynamics} \\ 
    & \mathbf{y}_k = h(\bar{\mathbf{x}}_k, \bar{\mathbf{u}}_k), \; k = \{0,\dots,N\}, \label{subeq:outputMapping} \\
    & \underline{\gamma} \leq \mathbf{u}_k \leq \bar{\gamma}, \; k = \{0,\dots,N-1\}, \label{subeq:actuationConstraints} \\
    &\dot{\underline{\gamma}} \leq \bar{\mathbf{u}}_k \leq \dot{\bar{\gamma}}, \; k = \{0,\dots,N\}, \label{subeq:accActuationConstraints} \\
    & \underline{d}_\mathcal{C} \leq \| \mathbf{d}_{\mathcal{C}_k} \| \leq \overline{d}_\mathcal{C}, \; k = \{ 0,\dots,N\}, \label{subeq:opticalLinkConstraint} \\
    & \frac{\mathbf{z}_T^\top (-\mathbf{d}_{\mathcal{C}_k})}{ \| \mathbf{d}_{\mathcal{C}_k} \|} \geq \cos{(\psi_\mathcal{C})}, \; k = \{ 0,\dots,N\}, \label{subeq:coneApertureConstraint} \\
    & \lVert \mathbf{p}_k - \mathbf{p}_{\mathcal{O}_{j,k}} \rVert^2 \geq d_{\mathcal{O}_j} + d_\mathrm{safe} - \varepsilon_{j,k}, \label{subeq:obstacleAvoidance} \\ 
    & \varepsilon_{j,k} \geq 0, \; j = \{1,\dots,N_{\mathcal{O}}\}, \nonumber 
    \end{align} 
    \normalsize
\end{subequations}
where~\eqref{subeq:objectiveFunction} defines the cost function, which minimizes a weighted sum of the output tracking error, the control rate, and the slack variables associated with obstacle avoidance. The output vector $\mathbf{y}_k$ includes position, velocity, acceleration, beam alignment metrics, and communication distance; it is compared against the desired output $\mathbf{y}_{d,k}$, with weighting matrix $\mathbf{Q}$ assigning relative importance to each component. The term $\lVert \bar{\mathbf{u}}_k \rVert^2_{\mathbf{Q}_{\bar{\mathbf{u}}}}$ penalizes rapid variations in control input, encouraging smoother actuation, while $\lVert \bm{\varepsilon}_k \rVert^2_{\mathbf{Q}_{\bm{\varepsilon}}}$ penalizes proximity to obstacles via slack variables.
Constraint~\eqref{subeq:initialCondition} sets the initial condition of the augmented state, while~\eqref{subeq:systemDynamics} and~\eqref{subeq:outputMapping} define system dynamics and output relations. Actuator constraints are captured by~\eqref{subeq:actuationConstraints} and~\eqref{subeq:accActuationConstraints}, enforcing physical bounds on rotor speeds and their rates of change.
The optical communication constraints are encoded in~\eqref{subeq:opticalLinkConstraint} and~\eqref{subeq:coneApertureConstraint}. These enforce that the \ac{MRAV} remains within the valid communication region: the former maintains distance bounds that ensure signal strength without attenuation or saturation; the latter ensures angular alignment between the beam direction and the receiver's acceptance cone, necessary for sustaining the link.
Obstacle avoidance is managed through the soft constraint~\eqref{subeq:obstacleAvoidance}. Each obstacle is represented as a spherical region with an added safety margin $d_\mathrm{safe}$. Slack variables $\varepsilon_{j,k} \geq 0$ allow temporary and penalized violations of this margin to retain problem feasibility in congested or dynamic environments. The severity of such violations is regulated by the penalty matrix $\mathbf{Q}_{\bm{\varepsilon}}$.
Together, this \ac{NMPC} formulation unifies tracking, actuation, communication, and safety constraints within a predictive optimization framework. It enables anticipatory and constraint-aware behavior suitable for aerial inspection tasks in cluttered, communication-critical environments.



\section{Simulation Results}
\label{sec:simulationResults}

To validate the proposed control framework, numerical simulations were performed using MATLAB and the MATMPC toolbox\footnote{\url{https://github.com/chenyutao36/MATMPC}}. The nonlinear optimal control problem was discretized using a fixed-step fourth-order Runge-Kutta integrator with sampling time $T_s = \SI{15}{\milli\second}$, and solved using the qpOASES solver\footnote{\url{https://github.com/coin-or/qpOASES}}. All simulations run on a laptop equipped with an Intel Core i7-8565U CPU at $\SI{1.80}{\giga\hertz}$ and $\SI{32}{\giga\byte}$ of RAM, running Ubuntu 20.04. Supplementary simulation videos are available at \url{https://mrs.fel.cvut.cz/nmpc-optical-comm}.

The full control architecture includes a reference generator running at $\SI{200}{\hertz}$, which provides the desired output trajectory $\mathbf{y}_d$, as well as the current position and velocity of the \ac{UGV}. These signals are fed into the  \acs{NMPC} controller, which runs at $\SI{500}{\hertz}$ and computes optimal rotor speeds $\bm{\Omega} \in \mathbb{R}^{N_p}_{\geq 0}$. The control inputs are applied to the \ac{GTMR} model simulated at $\SI{1}{\kilo\hertz}$, ensuring accurate representation of fast dynamics. The prediction horizon is to $\SI{0.75}{\second}$, discretized into $N=50$ steps. All relevant parameters are listed in Table~\ref{tab:controlParameters}.

\begin{table}[tb]
    \centering
    \caption{Optimization parameters.}
    \vspace{-0.5em}
    \label{tab:controlParameters}
    \begin{adjustbox}{max width=0.98\columnwidth}
    \begin{tabular}{c|c|c|c}
        \toprule
        \textbf{Sym.} & \textbf{Value} & \textbf{Sym.} & \textbf{Value} \\
        \midrule
        $\psi_\mathcal{C}$ & $0.17\,\si{\radian}$ & $\Upsilon$ & $1.0\,\si{\meter}$ \\
        $\underline{d}_\mathcal{C}$ & $0.25\,\si{\meter}$ & $\overline{d}_\mathcal{C}$ & $1.4\,\si{\meter}$ \\
        $\mathbf{p}(t_0)$ & $[-3.25; -3.25; 1]^\top\,\si{\meter}$ & $\mathbf{p}_R(t_0)$ & $[-3; -3; 0]^\top\,\si{\meter}$ \\
        $c_\xi$, $c_\tau$ & $1.18\!\times\!10^{-3}$, $2.5\!\times\!10^{-5}$ & $\mathbf{J}$ & $[0.11; 0.11; 0.19]^\top\,\si{\kilogram\meter\squared}$ \\
        $T_s$, $N$ & $15\,\si{\milli\second}$, $50$ & $\{\alpha, \beta\}$ & $\{20, 0\}^\circ$ \\
        $g$, $m$ & $9.81\,\si{\meter\per\square\second}$, $2.57\,\si{\kilogram}$ & $\bar{\gamma}$, $\underline{\gamma}$ & $100$, $16\,\si{\hertz}$ \\
        $\dot{\bar{\gamma}}$, $\dot{\underline{\gamma}}$ & $400$, $-200\,\si{\hertz\per\second}$ & $\mathbf{p}_\mathrm{BT}$ & $[0.1; 0; 0]^\top\,\si{\meter}$ \\
        $\psi_\mathcal{C}$, $\Phi_{1/2}$ & $89^\circ$, $10^\circ$ & $T_\mathrm{rx}$ & $26\,\si{\second}$ \\
        $d_\mathrm{safe}$, $d_\mathcal{O}$ & $0.25\,\si{\meter}$ & $N_\mathcal{O}$ & $3$ \\
        $\mathbf{p}_{\mathcal{O}_{1\!:\!3}}(t_0)$ & $[1.5\!; \!-3\!; 0.75]$, $[5\!; 1\!; 2]$, $[-2\!; 1.5\!; 0.5]$ & $\mathbf{p}_{\mathcal{O}_{1\!:\!3}}(t_N)$ & $[1.5\!; \!-3\!; 0.75]$, $[2\!; \!-1\!; 0.5]$, $[0\!; \!-3\!; 2]$ \\
        $\bar{\mathbf{Q}}$ & $\mathrm{diag}(0, 0.1, 0.1)$ & $\mathbf{Q}$ & $[\bar{\mathbf{Q}}; \mathrm{diag}(10, 10, 2)]^\top$ \\
        $\mathbf{Q}_{\bar{\mathbf{u}}}$ & $\mathrm{diag}(10,\dots,10)$ & $\mathbf{Q}_{\bm{\varepsilon}}$ & $\mathrm{diag}(10^4,\dots,10^4)$ \\
        \bottomrule
    \end{tabular}
    \end{adjustbox}
    \vspace*{-1.95em}
\end{table}

A representative test case for power infrastructure inspection was designed, featuring a bounded workspace of size $\{\SI{5}{\meter} \times \SI{5}{\meter} \times \SI{2}{\meter}\}$, with a simplified power tower at the center. The \ac{UGV} follows a square trajectory around the tower, relaying data to a remote base station via \ac{RF}. A \ac{MRAV}, modeled as a tilted \ac{GTMR} with nonzero rotor tilt angles $\{\alpha_i, \beta_i \}$ and $N_p = 6$, follows the \ac{UGV} while maintaining optical alignment. The transmitter is rigidly mounted $\SI{0.1}{\meter}$ ahead of the body frame origin.

The environment includes three obstacles: one static and two dynamic. The static obstacle $\mathbf{p}_{\mathcal{O}_1}$ remains fixed throughout the simulation, while the dynamic ones $\mathbf{p}_{\mathcal{O}_2}$ and $\mathbf{p}_{\mathcal{O}_3}$ follow straight-line trajectories. All obstacles are modeled as spheres with known positions over the horizon. The mission duration is $\SI{26}{\second}$, during which the \ac{MRAV} must track the \ac{UGV}, preserve optical link quality, and avoid collisions.

\begin{figure}[tb]
    \centering
    \input{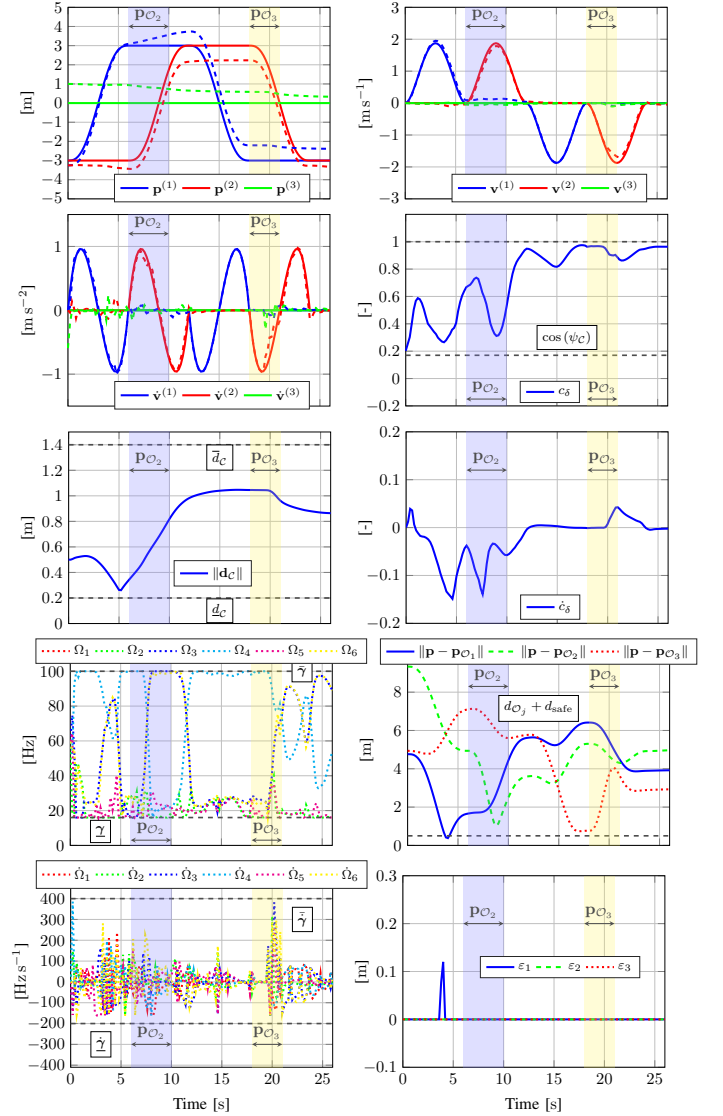}
    \vspace{-0.5em}
    \caption{Time evolution of constrained variables in the \acs{NMPC} problem. Solid lines: measured values; dashed: references. Superscripts $\bullet^{(1)}$, $\bullet^{(2)}$, and $\bullet^{(3)}$ denote $x$-, $y$-, and $z$-axis components. Shaded areas indicate motion of dynamic obstacles $\mathbf{p}_{\mathcal{O}_2}$ and $\mathbf{p}_{\mathcal{O}_3}$.}
    \label{fig:constraints}
    \vspace*{-1.75em}
\end{figure}

\begin{figure*}[tb]
    \centering
    \adjincludegraphics[width=1.8\columnwidth, trim={{.075\width} {.2\height} {.070\width} {.195\height}}, clip]{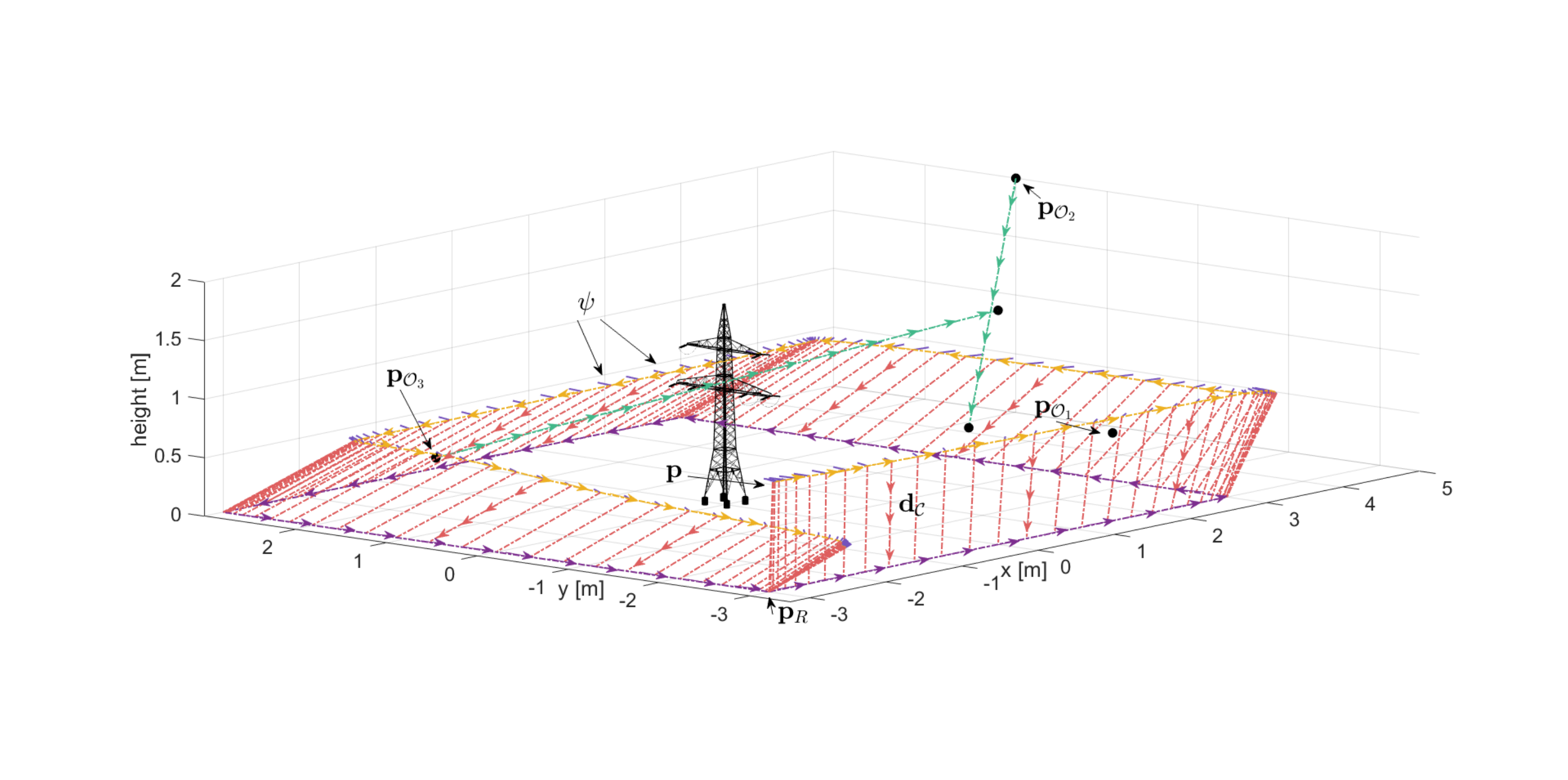}
    \vspace{-0.5em}
    \caption{Simulation overview showing \ac{MRAV} (yellow), \ac{UGV} (purple), and dynamic obstacles (green). Arrows indicate motion; drone heading (blue) and optical beam (red) are shown as body-fixed vectors.}
    \label{fig:fullScenario}
    \vspace*{-0.85em}
\end{figure*}

Figure~\ref{fig:fullScenario} illustrates the 3D environment, including the tower, obstacles, and trajectories of all agents. The beam vector (red) and drone heading (blue) are visualized, allowing inspection of optical alignment during flight.

\begin{figure}[tb]
    \centering
    \input{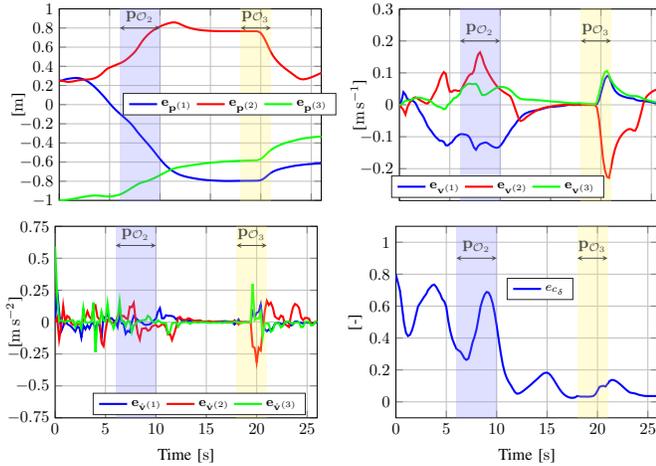}
    \vspace{-0.7em}
    \caption{Evolution of tracking errors for position, velocity, acceleration, and beam alignment. }
    \label{fig:plottingErrors}
\end{figure}

\begin{figure}[tb]
    \centering
    \vspace{-0.50em}
\hspace{-0.225cm}
\begin{subfigure}{0.425\columnwidth}
    \hspace*{-0.795cm}
    \centering
    \scalebox{0.62}{
    \begin{tikzpicture}
    \begin{axis}[%
    width=2.2119in,%
    height=1.6183in,%
    at={(0.758in,0.481in)},%
    scale only axis,%
    xmin=0,%
    xmax=26.0,%
    ymax=1.1,%
    ymin=-0.1,%
    xmajorgrids,%
    ymajorgrids,%
    ylabel style={yshift=-0.355cm, xshift=-0cm}, 
    xlabel={Time [\si{\second}]},%
    ylabel={[-]},%
    ytick={5,4,3,2,1,0,...,-5},%
    xtick={0,5,...,25},%
    axis background/.style={fill=white},%
    legend style={at={(0.65,0.33)},anchor=north,legend cell align=left,draw=none,legend 
    columns=-1,align=left,draw=white!15!black}
    ]
    \addplot [color=blue, solid, line width=1.25pt] 
        file{matlabPlots/constraints/I_rx_scope.txt};%
    \addplot [color=red, solid, line width=1.25pt] 
        file{matlabPlots/constraints/I_tx_scope.txt};%
    \fill[blue!50,nearly transparent] (axis cs:{6,-0.1}) -- (axis cs:{6,1.1}) -- (axis cs:{10,1.1}) -- (axis cs:{10,-0.1}) -- cycle;
    \fill[yellow!75,nearly transparent] (axis cs:{18,-0.1}) -- (axis cs:{18,1.1}) -- (axis cs:{21,1.1}) -- (axis cs:{21,-0.1}) -- cycle;
    %
    \draw[solid, stealth-stealth, black!70] (axis cs:6,0) -- node[below]{$\mathbf{p}_{\mathcal{O}_2}$} (axis cs:10.0,0);
    \draw[solid, stealth-stealth, black!70] (axis cs:18,0) -- node[below]{$\mathbf{p}_{\mathcal{O}_3}$} (axis cs:21.0,0);
    \legend{\footnotesize{$I_\mathrm{rx}$}, \footnotesize{$I_\mathrm{tx}$}};%
    %
    \end{axis}
    \end{tikzpicture}
    }
\end{subfigure}
\hspace*{-0.30cm}
\begin{subfigure}{0.425\columnwidth}
    \hspace*{0.095cm}
    \centering
    \scalebox{0.62}{
    \begin{tikzpicture}
    \begin{axis}[%
    width=2.2119in,%
    height=1.6183in,%
    at={(0.758in,0.481in)},%
    scale only axis,%
    xmin=0,%
    xmax=26.0,%
    ymax=1.1,%
    ymin=-0.1,%
    xmajorgrids,%
    ymajorgrids,%
    ylabel style={yshift=-0.355cm, xshift=0cm}, 
    xlabel={Time [\si{\second}]},%
    ylabel={[-]},%
    ytick={3,2,1,0,...,-3},%
    xtick={0,5,...,25},%
    axis background/.style={fill=white},%
    legend style={at={(0.55,0.33)},anchor=north,legend cell align=left,draw=none,legend 
    columns=-1,align=left,draw=white!15!black}
    ]
    \addplot [color=blue, solid, line width=1.25pt] 
        file{matlabPlots/constraints/I_scope.txt};%
    \fill[blue!50,nearly transparent] (axis cs:{6,-0.1}) -- (axis cs:{6,1.1}) -- (axis cs:{10,1.1}) -- (axis cs:{10,-0.1}) -- cycle;
    \fill[yellow!75,nearly transparent] (axis cs:{18,-0.1}) -- (axis cs:{18,1.1}) -- (axis cs:{21,1.1}) -- (axis cs:{21,-0.1}) -- cycle;
    %
    \draw[solid, stealth-stealth, black!70] (axis cs:6,0) -- node[below]{$\mathbf{p}_{\mathcal{O}_2}$} (axis cs:10.0,0);
    \draw[solid, stealth-stealth, black!70] (axis cs:18,0) -- node[below]{$\mathbf{p}_{\mathcal{O}_3}$} (axis cs:21.0,0);
    \legend{\footnotesize{$I$}};%
    %
    \end{axis}
    \end{tikzpicture}
    }
\end{subfigure}
    \vspace{-0.7em}
    \caption{Temporal evolution of the binary connectivity indicators $I$, $I_\mathrm{tx}$, and $I_\mathrm{rx}$ over the mission duration.}
    \label{fig:binaryVariables}
    \vspace*{-1.5em}
\end{figure}
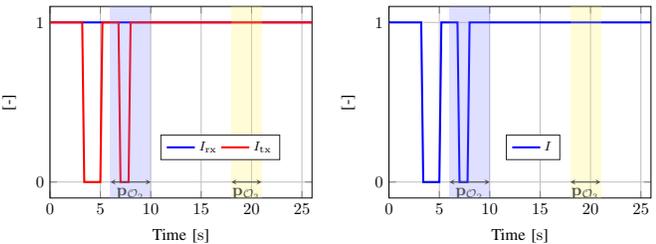

Figure~\ref{fig:constraints} plots critical variables subject to constraints in the \ac{NMPC} problem. Rotor speeds $\bm{\Omega}$ and their derivatives $\dot{\bm{\Omega}}$ remain within allowable bounds ($\underline{\gamma}$, $\bar{\gamma}$, $\dot{\underline{\gamma}}$, $\dot{\bar{\gamma}}$), demonstrating respect for actuation limits. The minimum distance to all obstacles respects the safety margin $d_\mathrm{safe}$. The inter-agent distance $\lVert \mathbf{d}_\mathcal{C} \rVert$ is kept within the optical link's viable range $[\underline{d}_\mathcal{C}, \overline{d}_\mathcal{C}]$, and the misalignment cosine $c_\delta$ remains above the threshold defined by $\psi_\mathcal{C}$, confirming that beam alignment is maintained. The figure also highlights a brief activation of the slack variable $\varepsilon_1$, indicating a temporary and minor violation of the minimum safe distance from obstacle $\mathcal{O}_1$. This behavior illustrates the ability of the controller to remain feasible even under tight spatial constraints, by leveraging penalized soft constraints.

Complementary performance metrics are shown in Fig.~\ref{fig:plottingErrors}, which reports tracking errors for position, velocity, and acceleration, as well as the evolution of $c_\delta$. These results confirm that the controller achieves high tracking accuracy while preserving precise optical pointing throughout the task.

In addition to constraint satisfaction, optical link performance was evaluated using the binary indicator $\bar{I} \in \{0,1\}$ and its time-averaged counterpart $\bar{I} \in [0,1]$, calculated over a sliding window of duration $T_\mathrm{rx} = \SI{26}{\second}$. At each time step, $I = 1$ if the transmitter is inside the receiver's cone and the distance constraints are satisfied. As shown in Figure~\ref{fig:binaryVariables}, link connectivity is maintained for most of the mission, yielding an average $\bar{I} = 0.8931$. This high value demonstrates the controller's ability to maintain persistent directional communication despite dynamic maneuvers, obstacle avoidance, and limited actuation authority.



\section{Conclusions}
\label{sec:conclusions}

This work presented a \acs{NMPC} framework for motion planning of \acp{MRAV} operating under directional optical communication constraints. The controller was designed to track a mobile ground robot equipped with optical receivers, enforcing strict constraints on beam alignment, transmitter–receiver distance, and obstacle avoidance while accounting for the full nonlinear dynamics and actuation limits of the aerial platform.
By embedding optical connectivity requirements into the \ac{NMPC} problem as hard constraints, the proposed approach enables safe, feasible, and communication-aware trajectories in structured inspection tasks. Simulation results confirmed the method's ability to maintain reliable \ac{FSO} communication, even in the presence of dynamic obstacles and actuation saturation, while ensuring tight tracking performance.
Future work will address connectivity loss and infeasibility due to strict beam or obstacle constraints by incorporating fallback strategies for safety. Environmental effects such as light interference, occlusions, wind, temperature, humidity, and rain will be modeled and tested via simulation. Finally, real-world experiments with \ac{MRAV}–\ac{UGV} systems will validate robustness, sensor uncertainty handling, and real-time performance.




\bibliographystyle{IEEEtran}
\bibliography{references}

\end{document}